\documentclass[a4paper]{article}
\usepackage{Odyssey2024}
\usepackage{epsfig,amssymb,amsmath,appendix,csquotes,xcolor}
\usepackage{float}
\ninept

\title{Adapting WavLM for Speech Emotion Recognition}

\name{Daria Diatlova$^{1,2,3}$, Anton Udalov$^{3,4}$, Vitalii Shutov$^{1,2}$, Egor Spirin$^{1,2,3}$}

\address{deepvk$^1$ , VK$^2$, VK Lab $^3$, ITMO$^4$\\
{\small \tt d.dyatlova@vk.team, a.udalov@niuitmo.ru, vi.shutov@corp.vk.com, egor.spirin@vk.team} }
\begin{document}
\maketitle

\begin{abstract}
Recently, the usage of speech self-supervised models (SSL) for downstream tasks has been drawing a lot of attention.
While large pre-trained models commonly outperform smaller models trained from scratch, questions regarding the optimal fine-tuning strategies remain prevalent. 
In this paper, we explore the fine-tuning strategies of the WavLM Large model for the speech emotion recognition task on the MSP Podcast Corpus.
More specifically, we perform a series of experiments focusing on using gender and semantic information from utterances.
We then sum up our findings and describe the final model we used for submission to Speech Emotion Recognition Challenge 2024. 
% By thoroughly exploring these facets, we aim to highlight the most important parameters and techniques that are worth experimenting with while fine-tuning the SSL model for the speech emotion recognition task.
\end{abstract}

\section{Introduction} \label{ssec:intro}
Speech Emotion Recognition (SER) is a highly-demanded task in practical applications. These include emotion recognition for analyzing customer satisfaction from voice messages and calls~\cite{vaudable2012negative}, creating a robust Automatic Speech Recognition (ASR) model~\cite{vegesna2019application}, or providing essential content for generating avatars in video calls~\cite{pan2024expressive}.

Recently, the interest among machine learning researchers in using self-supervised learning (SSL) models for audio processing has grown~\cite{NEURIPS2020_92d1e1eb,pmlr-v162-baevski22a,chen2021wavlm}. 
This can be explained by the successful application of transfer learning, which involves leveraging the knowledge of SSL upstream models, initially trained on a large amount of unlabeled data, by a downstream model~\cite{iman2023review}. 
The most commonly used method to transfer knowledge to a downstream model involves fine-tuning the upstream model by adding new layers from the downstream model~\cite{iman2023review}.

The application of transfer learning in the audio domain, particularly SER, has received considerable attention of late.
In~\cite{boigne_2020}, the authors demonstrated that using an SSL model as a feature extractor outperforms SER solutions based on low-level descriptors, even when provided with eight times less labeled data. 
Following that, works focusing on the specific utilization of SSL to enhance SER performance have emerged~\cite{boigne_2020,xia21b_interspeech,morais2022speech}. 
These approaches include feature-wise objective constraints~\cite{xia21b_interspeech}, the utilization of text encoders~\cite{boigne_2020,morais2022speech}, pooling mechanics~\cite{kakouros2022speechbased}, and more.

In this paper, we share the technical details of creating a SSL-based model for speech emotion recognition using WavLM~\cite{chen2021wavlm} in the Odyssey 2024 challenge~\cite{Goncalves_2024}, and address several research questions. 
\begin{itemize}
\item Does applying time-dimensional pooling over WavLM output impact the quality of speech emotion recognition?
\item Can incorporating information about the speaker's gender enhance emotion classification?
\item Does leveraging information about the text of an utterance help emotion classification?
\end{itemize}
% Based on the questions posed, we formulated several hypotheses on how WavLM fine-tuning for SER can be improved.
Using the MSP Podcast dataset for verification, we obtained a solution which we then used for the Odyssey 2024 challenge.

% continue the exploration of SSL fine-tuning approaches for constructing a robust model for speech emotion recognition. Specifically, we share our findings from fine-tuning the WavLM model for speech emotion recognition using the MSP Podcast dataset during the Odyssey 2024 challenge. 
% The proposed model consists of fusing 5 models described later in the paper. 
% The main contribution of this work is that, by accumulating the results from our experiments, we share the details of the trained model for Odyssey 2024 challenge participation. 
% This includes the details of semantic and gender information accumulation, supported by the ablation study.

\section{Related Work}
In this section, we discuss WavLM~\cite{chen2021wavlm}, a self-supervised audio processing model, and its implications for SER. 
We explore cutting-edge strategies for information gathering and highlight the importance of advanced pooling techniques. Additionally, we consider the incorporation of auxiliary data such as text and gender embeddings to enhance SER systems.

\subsection{WavLM} \label{ssec:wavlm-related}
WavLM~\cite{chen2021wavlm} is a self-supervised audio processing model that utilizes the Transformer encoder architecture~\cite{vaswani2017attention}. 
The model has been enhanced with a gated relative position bias, improving its ability to model sequential information. 
Specifically, WavLM takes a raw audio signal $X \in \mathbb{R}^n$, where $n$ is the number of samples in the raw signal, and produces an encoded output $Z \in \mathbb{R}^{[l \times m \times h]}$, where: 
\begin{itemize}
    \item $l$ is the number of Transformer layers,
    \item $m$ is the sequence length,
    \item $h$ is the hidden dimension. 
\end{itemize}
To accommodate various computational needs and applications, WavLM is available in two configurations: Base, with 12 Transformer layers, and Large, with 24 layers. 
The WavLM model is trained on extensive audio corpora to encode a wide range of audio applications. 
It incorporates a speech denoising objective and masked speech prediction during pre-training, enabling efficient extraction of speech and acoustic features.

\subsection{Self-Supervised Learning in SER} \label{sssec:related:ssl}
Self-supervised learning (SSL) has shown promising results in SER~\cite{boigne_2020,xia21b_interspeech}, offering innovative ways to decode emotional subtleties in speech. 
% The integration of wav2vec 2.0 and HuBERT for upstream processing, as detailed in~\cite{lashkarashvili2024parameter}, illustrates the utility of pre-trained models in capturing a broad spectrum of emotional attributes efficiently. This framework demonstrates how fine-tuning with emotion-specific attributes enhances domain adaptability and recognition accuracy.
It is common for SER models built using SSL to use the Upstream + Downstream model architecture~\cite{morais2022speech}. 
The upstream block aims to extract features for the downstream model to predict emotion classes.

Before passing the upstream model's output to the downstream model, it is typically reduced in both the layer and time dimensions.
A common approach for layer-wise reduction is: 
\begin{itemize}
    \item to take the $\hat{Z} \in \mathbb{R}^{[l_n \times m \times h]}$ hidden representation, where $l_n$ is the output from the last Transformer layer~\cite{chen2024vesper}.
    \item to select the weighted average sum of all layers~\cite{chen2024vesper}.    
\end{itemize}
For the second case, a vector of trainable weight parameters $w \in \mathbb{R}^l$ is initialized.
Then, a layer-wise averaged output $\hat{Z} = Z \cdot w$, where $\hat{Z} \in \mathbb{R}^{[m \times h]}$, is passed through the time dimension reduction block.
The simplest implementation of the time-reduction block is to apply average pooling~\cite{krizhevsky2012imagenet}: $x_{\text{out}} = \sum_{i=1}^m \frac{\hat{Z}_i}{m}$, where $x_{\text{out}} \in \mathbb{R}^{[h]}$.
After this step, $x_{\text{out}}$ is passed to the downstream block for further classification.

The choice of the downstream architecture, as well as the modification applied to the process of obtaining $x_{\text{out}}$, can enhance the quality of the SER model.

The combination of mean average pooling and sophisticated aggregators, such as ECAPA-TDNN~\cite{desplanques2020ecapatdnn}, with linear classifiers improved SER performance in~\cite{morais2022speech}.
In~\cite{kakouros2022speechbased}, attentive correlation pooling by adding weights to the statistical estimates was introduced. 
This technique aims to address the challenge of utilizing emotional information across extended periods of time.
Additionally, the authors of~\cite{kakouros2022speechbased} demonstrated the benefits of label smoothing for dealing with the emotional ambiguity of speech.
State-of-the-art SER performance was achieved by combining attentive correlation pooling with label smoothing and dropout, without the need for fine-tuning the SSL model.

\subsection{Information Accumulation}
Information accumulation techniques, particularly those involving pooling mechanisms and the integration of auxiliary information such as text and gender embeddings, play a pivotal role in enhancing SER systems.

In~\cite{gao23d_interspeech}, introduced a two-stage fine-tuning process for wav2vec 2.0, emphasizing the use of automatic speech recognition (ASR) and gender pretraining to enrich the SER model with additional features. 
By first embedding these attributes into the model and then focusing on emotion recognition, the approach effectively leverages multitask learning (MTL) and adversarial learning to incorporate auxiliary task information, aiming to mitigate the gradient conflict issue commonly observed in MTL setups.
This method highlights the utility of informed feature extraction for boosting SER performance.

Furthermore, the concept of attentive statistics pooling~\cite{okabe2018attentive} proposes an advanced pooling strategy that goes beyond the traditional average pooling method by including higher-order statistics, such as standard deviation, weighted by attention mechanisms. 
This approach calculates both the weighted mean and standard deviation of frame-level features, enhancing the model's ability to capture speaker discriminability through feature variability over time. By applying attention to both means and variances, it introduces a more nuanced understanding of speaker characteristics vital for SER.

\section{Model Modifications}  \label{sssec:proposed_model}
In this section, we propose several modifications aimed at addressing our research questions. More specifically, we describe 1) replacing the commonly used average pooling with pooling types, 2) using gender information for predictions, and 3) using text information in conjunction with the pooled output.

\subsection{Time-Dimensional Pooling} \label{sssec:proposed_model_pooling}

The output of WavLM is usually reduced over time-dimension using simple average pooling and then passed to the classification head, as discussed in Section~\ref{sssec:related:ssl}.
Instead of average pooling, we suggest using STD and attention pooling.

\subsubsection{STD Pooling}
In standard deviation (STD) pooling the average pooling output (Eq. \ref{eq1}, where $m$ is the sequence length and $Z_i$ is the layer-wise averaged output of WavLM for a time frame $i$) is used along with the standard deviation output (Eq. \ref{eq2}). 
The resulting statistics are concatenated over the hidden dimension (Eq. \ref{eq3}).
Compared to a simple average pooling, which is generally robust to noisy output, STD pooling focuses on regions with high variability. 
It can emphasize distinct features that differ from neighboring regions, which can be beneficial for emotion classification.

\begin{equation}
x_{\text{mean}} = \sum_{i=1}^m \frac{\hat{Z}_i}{m}
\label{eq1}
\end{equation}

\begin{equation}
x_{\text{std}} = \sqrt{\frac{\sum_{i=1}^m (\hat{Z}_i - x_{\text{mean}})^2}{m}}
\label{eq2}
\end{equation}
\begin{equation}
x_{\text{pooled}} = [x_{\text{mean}}, x_{\text{std}}] 
\label{eq3}
\end{equation}

\subsubsection{Attention Pooling} \label{sssec:proposed-attention-pooling}
Attention pooling~\cite{okabe2018attentive} makes it possible for certain frames to become more valuable based on context.
Attention weights are computed as shown in Eq.~\ref{eq4}, where $p \in \mathbb{R}^{h}$ is the trainable parameter and $\cdot$ denotes matrix multiplication.

$\hat{Z}$ is multiplied element-wise by the attention weights (Eq.~\ref{eq5}), then the mean and variance are computed from the resulting output and concatenated as described in Eq.~\ref{eq3}.

\begin{equation}
w_{\text{att}} = \text{Softmax}(p \cdot \tanh(\hat{Z})), \space w_{\text{att}} \in \mathbb{R}^m
\label{eq4}
\end{equation}
\begin{equation}
\hat{Z} = \hat{Z} \space \cdot \space w_{\text{att}}, \space \hat{Z} \in \mathbb{R}^{[m \times h]}
\label{eq5}
\end{equation}

Attention pooling provides the opportunity to give certain frames more weight than others when computing the mean and standard deviation.

\subsection{Gender Conditioning}
\label{sssec:gender-conditioning}

The MSP Podcast corpus includes gender labels: male and female. 
Therefore, we added a lookup table of trainable embeddings. 

\begin{equation}
x_{\text{output}} = x_{\text{pooled}} \cdot e_{\text{gender}}
\label{eq9}
\end{equation}
For the conditioning mechanism, we used a pointwise  multiplication of pooled output and gender embeddings as shown in Eq. \ref{eq9}. 
In section \ref{sssec:gender-conditioning-exp}, we provide an ablation study on other conditioning mechanisms. 

\subsection{Text Conditioning} \label{sssec:text-conditioning}
Along with gender information, the MSP Podcast corpus contains textual annotations paired with audio clips. 
To combine the textual information with the encoded audio representation and gender embeddings, we employed the Sentence Transformer \cite{reimers-2020-multilingual-sentence-bert}, which maps transcriptions to a 384-dimensional dense vector, $e_{\text{text}}$. Conditioning was applied as depicted in Eq. \ref{eq6}, where $F$ represents a sequence of transformations: Linear Layer~\cite{rumelhart1986learning}, Layer Normalization~\cite{ba2016layer}, ReLU~\cite{agarap2018deep}, and Dropout~\cite{hinton2012improving}.

\begin{equation}
x_{\text{out}} = \frac{x_{\text{pooled}} + e_{\text{gender}} + F(e_{\text{text}})}{3}
\label{eq6}
\end{equation}

The ablation study on conditioning approaches can be found in Section \ref{sssec:exp_text}.

\begin{figure*}[t]
    \centering
    \includegraphics[width=1.0\textwidth]{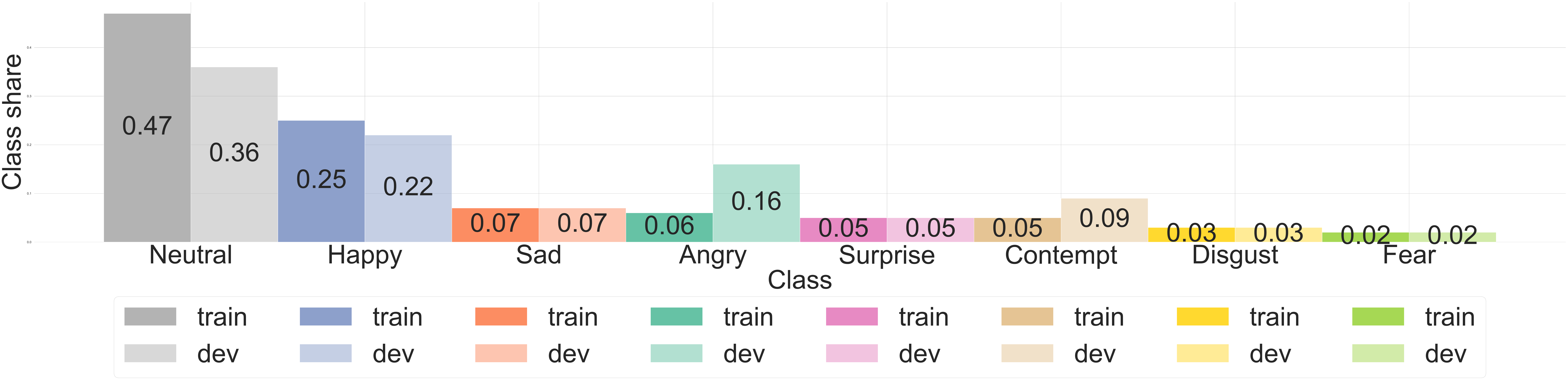}
    \caption{{\it Label distribution in the training and development sets. Note that the dataset is highly imbalanced, and the distribution of samples for each class relative to the total number of classes differs between the training and validation datasets.}}
    \label{train_dist}
\end{figure*}

\section{Experimental Setup}
In this section, we provide an overview of the dataset, the baseline proposed by the Odyssey Challenge 2024~\cite{Goncalves_2024} Organizers, and the setup for training and evaluating the models.
\subsection{Dataset}
The dataset used for training and evaluation is the MSP Podcast Corpus~\cite{martinez2020msp}. 
It consists of 90,522 speaking turns, with the original train, development, and test3 split. 
The test3 dataset does not have publicly available labels. 
Therefore, all our experiments, except for the final submission result, are presented using only the development set.

For the train and development set labels, we used the consensus (plurality vote of 3 out of 5 assessors) of the primary emotion, filtering out the classes \enquote{Other} and \enquote{No agreement}. 
This resulted in 53,386 samples in the training set and 15,341 samples in the development set.
On the early stages of experiments we collected the \enquote{confident} training set by taking the consensus (plurality vote of 4 out of 5 assessors) of the primary emotion. 
This resulted in just 23,509 samples, but did not lead to any improvement of the F1-macro score. 

The MSP Podcast dataset exhibits a high level of imbalance, with the class distribution of the training and development sets shown in Figure~\ref{train_dist}.
Note that the order of the number of samples of each class relative to the total number of classes is different in the training and validation datasets.
Early on in our experiments, we attempted to balance the dataset by augmenting utterances from classes with fewer samples, but did not observe any significant improvements.

The test3 dataset lacks gender labels, unlike the training and validation datasets.
To obtain gender labels, we utilized an internal gender-prediction model. 
As for text transcriptions, we observed that some examples in the training and validation datasets also lack text transcriptions. 
We used the Whisper~\cite{radford2022robust} V3 Large model to transcribe these audio recordings, along with the test3 dataset.

\subsection{Baseline} \label{sssec:baseline}
An official baseline was provided by the organizers of the Odyssey 2024 challenge~\cite{Goncalves_2024}.
The baseline model is a WavLM Large model, as described in Section~\ref{ssec:wavlm-related}, which produces the output $Z \in \mathbb{R}^{[l \times m \times h]}$, where $l=24$ (Transformer layers),  $h=768$ (hidden dimension), and $m$ is the dynamically changed sequence length shape.

For classification, the weighted average sum of all layers is selected as discussed in Section~\ref{sssec:related:ssl}. 
The pooled output is then passed through the projection layer, which consists of the Linear Layer~\cite{rumelhart1986learning}, Layer Normalization~\cite{ba2016layer}, ReLU~\cite{agarap2018deep}, and Dropout~\cite{hinton2012improving}.
Then, the attention pooling block (Section~\ref{sssec:proposed-attention-pooling}) is applied to the output and passed through the final Linear Layer to get the logits for each emotion class.

The model is trained with the weighted cross entropy loss~\cite{cox1958regression}, the weights are declared as shown in Eq.~\ref{eq11}, where $w_i$ is the weight for class $i$, $N_{\text{total}}$ is the size of train dataset, $m$ is the number of classes and $N_{\text{class}}$ is the number of samples in class $i$.

\vspace*{-5mm}
\begin{equation}
w_i = \frac{N_{\text{total}} \cdot N_{\text{class}}}{m}
\label{eq11}
\end{equation}

\subsection{Training and Evaluation}
All our experiments are based on WavLM Large. Unlike the baseline model, we used a single Linear Layer as a projection layer. 
We did not notice a significant difference in quality from the baseline model. However, our version is less memory-intensive, which is crucial when training large models.

The size of the projector linear layer is set to 256 by default unless specified otherwise. We performed various experiments by adjusting the projection layer size, with the results detailed in Table \ref{projection}. 
From the results, it is evident that the optimal choices for the projection layer size are 16, 32, and 256. 
The default selection of a hidden size of 256 is motivated by the requirement to combine the output of the projection layer with gender and text embeddings, where a larger embedding size makes it possible to encode more information.

In some of our experiments, we apply label smoothing within cross-entropy, which is motivated by emotional data ambiguity.
The transition from hard labels to soft labels consistently outperforms cross-entropy loss and leads to faster convergence during training~\cite{zhou2022all}. 
Label smoothing is defined by parameter $\gamma$, where $1-\gamma$ of the weight is assigned to the true label, and $\gamma$ is distributed across other classes.

Table~\ref{smoothing} presents the results from applying label smoothing.
It can be seen that, in our case, the usage of label smoothing leads to F1-macro score degradation.

All models were trained using the AdamW optimizer \cite{DBLP:conf/iclr/LoshchilovH19} with a learning rate of 1e-5 and weight decay of 0.01 for 20 epochs employing a batch size of 32 on a single A100 GPU. 

We keep the layers of WavLM CNN feature encoder frozen during all experiments.
For model training, we use audio clips of no longer than 5 seconds. 
For longer utterances, a 5-second segment is randomly selected on each epoch using cropping.

\begin{table}[ht]
\caption{\label{table:proj_size} {\it Best F1-macro scores on the development set with varying pooling output size.}}
\vspace{2mm}
\centerline{
\begin{tabular}{|c|c|}
\hline
Projection layer size & F1-macro \\
\hline  \hline
16 & \textbf{0.31} \\
32 & \textbf{0.31} \\
64 & 0.28 \\
128 & 0.30 \\
256 & \textbf{0.31} \\
\hline
\end{tabular}
}
\label{projection}
\end{table}

\begin{table}[th]
\caption{\label{table:label_smoothing} {\it Best F1-macro scores on the development set with varying portion of label smoothing.}}
\vspace{2mm}
\centerline{
\begin{tabular}{|c|c|c|}
\hline
Label smoothing & F1-macro \\
\hline  \hline
0.0  & \textbf{0.31} \\
0.1 & 0.30 \\
0.2 & 0.28 \\
\hline
\end{tabular}
}
\label{smoothing}
\end{table}

\section{Experiments} \label{ssec:experiments}
In this section, we answer the research questions stated in Section~\ref{ssec:intro} by conducting our experiments with the modifications from Section~\ref{sssec:proposed_model}.
For statistical significance, we run each experiment with several seeds and average the results.

\subsection{Does applying time-dimensional pooling over WavLM output impact the quality of speech emotion recognition?}
We conducted experiments using STD and attention pooling as described in Section~\ref{sssec:proposed_model_pooling}, and compared the results to using standard average pooling. 

From the results presented in Table~\ref{pooling}, it is evident that the change of pooling type influences overall performance of the SER model based on F1-macro score. 
STD pooling yielded the best F1-macro score in our experimental setup, while attention pooling performed worse than standard average pooling in our experiment setup.

Note that the baseline model, which uses average pooling, achieves an F1-macro score of 0.31 on the development set. 
Therefore, it is worthwhile to increase the number of training steps to achieve its best performance.
However, we chose to use STD pooling for the rest of our experiments as it converges faster, which is crucial for working within the Challenge.

\begin{table}[th]
\caption{\label{table:pooling} {\it Best F1-macro scores on the development set with varying temporal pooling type during training.}}
\vspace{2mm}
\centerline{
\begin{tabular}{|c|c|}
\hline
Pooling type & F1-macro \\
%& F1 A & F1 H & F1 N & F1 S & F1 C & F1 U & F1 F & F1 D 
\hline  \hline
average & 0.30 \\
std & \textbf{0.31} \\
attention & 0.28 \\
\hline
\end{tabular}
}
\label{pooling}
\end{table}

\subsection{Can incorporating information about the speaker's gender enhance emotion classification?} \label{sssec:gender-conditioning-exp}
As outlined in Section~\ref{sssec:gender-conditioning}, we utilized gender information to condition the pooled output. 
For this ablation study, we employed STD temporal pooling.

In this setup, \enquote{sum} and \enquote{multiplication} denote the addition and the multiplication of the pooled output with the gender embedding, \enquote{sum / 2} indicates that the output is divided by 2 after addition.
In the \enquote{stack + linear} configuration, the gender embedding and pooled output are concatenated along the hidden dimension and passed through a linear layer. 
The term \enquote{CLN} refers to conditioning layer normalization~\cite{ba2016layer}, as shown in Eq.~\ref{eq:cln}, where $\mathrm{g}$ and $\mathrm{f}$ represent linear layers.

\begin{equation}
  \label{eq:cln}
  y = \mathrm{f}(e_{\text{gender}})  \cdot \ \frac{x_{\text{out}} - \mathrm{mean}}{\mathrm{var}} + \mathrm{g}(e_{\text{gender}}),
\end{equation}

\vspace*{-5mm}
\begin{table}[th]
\caption{\label{gender} {\it Best F1-macro scores on the development set with varying gender conditioning types. Note that F1-macro for \enquote{sum} and \enquote{sum / 2} is the highest.}}
\vspace{2mm}
\centerline{
\begin{tabular}{|c|c|}
\hline
Conditioning & F1-macro \\
\hline  \hline
no conditioning & 0.31\\
sum & \textbf{0.32} \\
sum / 2 & \textbf{0.32} \\
multiplication & 0.31 \\
stack + linear & 0.29 \\
CLN & 0.31 \\
\hline
\end{tabular}
}
\end{table}

As shown by the results in Table~\ref{gender}, the sum and normalized sum reach the best F1-macro scores.
CLN and multiplication reach the second best F1-macro score, but it is equal to no conditioning case. 
Meanwhile, the stack + linear setup leads to F1-macro degradation.

In summary, combining the WavLM output with gender information can enhance the accuracy of the SER model, but the specific strategy of combining them is crucial.

\subsection{Does leveraging information about the text of an utterance help emotion classification?} \label{sssec:exp_text}
In this experiment, we followed a setup similar to the one in Section \ref{sssec:text-conditioning}, incorporating STD pooling together with gender and text embeddings.
The conditioning is applied using multiplication, normalized sum (\enquote{sum / 3}), and CLN.

For CLN, we first concatenated the gender and text embeddings along the hidden dimension and passed the resulting vector through a linear layer to reduce the hidden dimension of a conditioned vector to the initial size of 256.

From the results presented in Table~\ref{text}, it is evident that the addition of text information led to the degradation in the F1 score compared to using only the gender embedding. 
The best result was achieved using CLN and sum / 3 conditioning.

\begin{table}[th] \label{ssec:proposed_text}
\caption{\label{table:text_cond} {\it Best F1-macro scores on the development set with varying text conditioning types.}}
\vspace{2mm}
\centerline{
\begin{tabular}{|c|c|}
\hline
Conditioning & F1-macro \\
\hline  \hline
gender conditioning only & \textbf{0.32} \\
sum / 3 & 0.31 \\
multiplication & 0.30 \\
CLN & 0.31 \\
\hline
\end{tabular}
}
\label{text}
\end{table}

\begin{table*}[ht]
\centering
\caption{\label{table:conf} {\it The configurations of the models used for the submission. Note that out of the 5 models used for prediction fusion, three were conditioned on gender, and two were conditioned on text.}}
\label{model_configuration}
\begin{tabular}{|l|l|l|l|l|l|}
\hline
\textbf{Parameter} & \textbf{Model 1 (Baseline)} & \textbf{Model 2}  & \textbf{Model 3}  & \textbf{Model 4}  & \textbf{Model 5}  \\ \hline
Temporal pooling & attention & std & std & std & std \\
Pooled output size & 256 & 32 & 256 & 256 & 256                         \\
Gender conditioning & - & - & multiplication & sum / 3 & sum / 3 \\ 
Text conditioning & - & - & - & sum / 3 & sum / 3 \\ 
Label smoothing & 0.0 & 0.0 & 0.0 & 0.0 & 0.1 \\ 
\hline
\end{tabular}
\end{table*}

\begin{table*}[ht]
\centering
\caption{\label{table:models} {\it F1 scores for the models used in the challenge submission. Note that each model achieves the highest F1 score for several classes. The F1-macro score received after the fusion is higher then for other models.}}
\label{model_f1}
\begin{tabular}{|l|l|l|l|l|l|l|l|l|l|}
\hline
\textbf{Model} & \textbf{F1 Macro} & \textbf{Neutral}  & \textbf{Happy}  & \textbf{Angry}  & \textbf{Contempt} & \textbf{Sad} & \textbf{Surprise} & \textbf{Disgust} & \textbf{Fear} \\ \hline
Model 1 (Baseline) & 0.31 & 0.55 & 0.57 & 0.40   & 0.20 & 0.13 & \textbf{0.42} & 0.03 & \textbf{0.15} \\
Model 2 & 0.33 & \textbf{0.58} & 0.52 & \textbf{0.53} & 0.15 &0.40 & 0.19 & 0.21 & 0.04                        \\
Model 3 & 0.32 & 0.51 & 0.58 & \textbf{0.53} & \textbf{0.38} & 0.21 & 0.20 & 0.14 & 0.02\\ 
Model 4 & 0.31 & 0.13 & \textbf{0.59} & 0.40   & 0.15 & \textbf{0.42} & 0.21 & \textbf{0.55} & 0.04\\ 
Model 5 & 0.31 & 0.29 & \textbf{0.59} & 0.42 & 0.15 & \textbf{0.42} & 0.21 & 0.12 & 0.05\\ 
\hline
Fusion & \underline{0.35}  & 0.53  & 0.57 & \underline{0.61} & \underline{0.43} & 0.17 & 0.24 & 0.19 & 0.05\\ 
\hline
\end{tabular}
\end{table*}

In conclusion, combining the text information with the encoded WavLM output and gender embeddings did not improve the accuracy of the SER model in our specific case.
This can be explained by the chosen text encoder's insensitivity to emotional speech, as it encodes text primarily based on context. 
Alternatively, it may be due the limited correlation between text and emotion in most classes, or the strategies used to combine this information.

It is important to mention that we included text embedding in the WavLM output along with the information about gender in our experiment setup. 
This addition helped us enhance the quality of emotion recognition in the previous experiment (Section~\ref{sssec:gender-conditioning}). 
However, combining it with information about the text may negatively impact recognition quality, as it reduces the total information available in the resulting hidden vector about the WavLM output.

\section{Results}
In this section, we explain how we combined several models for our Odyssey 2024 challenge submission and discuss the final results.

\subsection{Fusion Process}
Fusing the predictions from multiple models improve the generalization ability and can lead to improved performance~\cite{dietterich2000ensemble}.
We employed the Constrained Optimization By Linear Approximation algorithm~\cite{nannicini2019performance} to train the weights matrix $w_{\text{fusion}} \in \mathbb{R}^{n_m \times l}$, where $n_m$ is the number of models and $l$ is the number of classes.
The predicted label $p$ from the predictions matrix $M \in \mathbb{R}^{n_m \times l}$ was determined as shown in Eq.~\ref{eq7}. The weights matrix $w_{fusion}$ was trained with the following constraints: $\sum_{i=1}^{n_m}  w_{fusion}^i = \sum_{i=1}^{l}w_{fusion}^i = 1$.

\vspace*{-2.5mm}
\begin{equation}
p = argmax(\sum_{i=1}^{n_m} M \cdot w_{fusion})
\label{eq7}
\end{equation}

\subsection{Model Configuration}
We aimed to combine the best of our models based on the F1-macro score and the models with the most diverse configurations.
As a result, we choose 5 models with configurations described in Table~\ref{model_configuration}. 
The F1 score for each emotion class and F1-macro for each model are reported in Table~\ref{model_f1}.

Please note that the results in Table~\ref{model_f1} were obtained during our work on the challenge, and there are some differences in the F1-macro compared to the results reported in Section~\ref{ssec:experiments}.
For instance, model 3 in Table~\ref{model_f1} has the same configuration as the model in the fourth row of Table~\ref{gender}, but it shows an F1-macro of 0.32, as opposed to 0.31. 
This difference is due to running additional experiments after submission to provide statistically significant results for the research questions in Section~\ref{ssec:experiments}. 

\subsection{Result Analysis}
Table~\ref{model_f1} demonstrates that each of the models selected for fusion achieves the highest F1 score for two or three emotional classes compared to the others. 
The idea behind fusion is to leverage the strengths of all models. 
The last row in Table~\ref{model_f1} displays the F1 scores for the fused model predictions. 
By combining predictions from the 5 models, we achieved an F1-macro score of 0.35, which is 0.02 higher than the best individual model's F1 score. 
Notably, the F1 scores for the \enquote{Angry} and \enquote{Contempt} classes after fusion are also higher than the maximum among the 5 models. 
For the other classes, the F1 score lies between the lowest and highest values in the column.

\section{Conclusion}
In this paper, we discussed the modifications made to the WavLM model during its fine-tuning for speech emotion recognition as part of our participation in the Odyssey Challenge 2024.
We posed several research questions to investigate the importance of time-dimensional pooling, gender, and text accumulation in training the SER model.
Our experiments on the MSP Podcast Corpus led us to the conclusion that incorporating STD pooling with accumulated gender information could enhance the performance of SER models based on WavLM.

\bibliographystyle{IEEEbib}
\bibliography{Odyssey2024_BibEntries}

\end{document}